%% file: main.tex
\documentclass[twocolumn]{article} % Or use 'IEEEtran' for IEEE format, etc.

\usepackage{multicol}  % For two-column support
\usepackage[letterpaper, top=0.6in, bottom=0.5in, inner=0.75in, outer=0.75in, headsep=0.3in, footskip=0.3in]{geometry}

\usepackage{titlesec}
\titleformat{\section}{\large\bfseries}{\thesection}{1em}{}
\titleformat{\subsection}{\normalsize\bfseries}{\thesubsection}{1em}{}

\usepackage{graphicx}         % For including images
\usepackage{amsmath, amssymb} % For math symbols and equations (even if you don't use much, good practice)
\usepackage{hyperref}          % For hyperlinks (URLs, references)
\usepackage{natbib}          % For bibliography management (BibTeX style)
\usepackage{geometry}         % For adjusting margins if needed
\usepackage{listings}
\usepackage{pgfplots}
\usepackage{amssymb} % For \checkmark
\usepackage{pifont}  % For \xmark
\usepgfplotslibrary{fillbetween}
\usepackage{algorithm}
\usepackage{algorithmic}
\usepackage{amsmath} 
\usepackage{appendix}

\title{\textbf{PoseLess: Depth-Free Vision-to-Joint Control via Direct Image Mapping with VLM}}

\author{
    Alan Dao (Gia Tuan Dao)\textsuperscript{1}, Dinh Bach Vu\textsuperscript{1} ,Tuan Le Duc Anh\textsuperscript{1}, Bui Quang Huy\textsuperscript{1} \\
    Menlo Research \\
    \texttt{alan@menlo.ai, bach@menlo.ai, charles@menlo.ai, yuuki@menlo.ai} \\
    \textsuperscript{1}Equal contribution.
}

\date{February 20, 2025} % Or specify a date, or use \date{} for no date

\begin{document}

\maketitle
\begin{abstract}
 % Prevents indenting the abstract paragraph
\noindent This paper introduces PoseLess, a novel framework for robot hand control that eliminates the need for explicit pose estimation by directly mapping 2D images to joint angles using projected representations. Our approach leverages synthetic training data generated through randomized joint configurations, enabling zero-shot generalization to real-world scenarios and cross-morphology transfer from robotic to human hands. By projecting visual inputs and employing a transformer-based decoder, PoseLess achieves robust, low-latency control while addressing challenges such as depth ambiguity and data scarcity. Experimental results demonstrate competitive performance in joint angle prediction accuracy without relying on any human-labelled dataset.
\end{abstract}

% --- Include Sections ---
\input{sections/1_introduction.tex}
\input{sections/2_related_work.tex}
\input{sections/3_methodology.tex}
\input{sections/4_experiments.tex}

\input{sections/5_discussion.tex}
\input{sections/6_conclusion.tex}

\bibliographystyle{plainnat} % Choose your bibliography style (e.g., 'plainnat', 'abbrvnat', etc.)
\bibliography{bibliography} % 'bibliography' is the name of your .bib file (without the extension)
% Note down some papers here:

% --- Appendices (Optional) --- 
% --- Appendices (Optional) ---

% --- Appendix A: Detailed Experimental Setup ---
% \clearpage
% \clearpage
% \appendix
% \onecolumn
% \input{sections/appendices.tex}
% ... Appendix content ...

\end{document}

%% file: sections/1_introduction.tex
\section{Introduction}

Robotic hand control has traditionally relied on explicit pose estimation to bridge the gap between visual perception and motor execution. This process typically involves extracting keypoints or 3D skeletal representations from images before translating them into joint commands. While effective, such approaches suffer from inherent limitations, including accumulated errors from multi-stage processing pipelines, sensitivity to occlusions, and dependency on high-quality labeled datasets. Furthermore, conventional methods often require careful calibration between the vision and actuation systems to maintain accuracy.

Recent advances in computer vision and deep learning have opened new possibilities for direct perception-to-control models that bypass intermediate pose estimation. By leveraging large-scale vision-language models (VLMs) with robust image projection strategies, it is now feasible to map high-dimensional image inputs directly to control outputs without requiring an explicit representation of pose. This paradigm shift enables more flexible and efficient control policies that are less susceptible to the limitations of traditional approaches.

\subsection{Motivation}

\label{sec:introduction}
\begin{figure}
    \centering
    \includegraphics[width=1\linewidth]{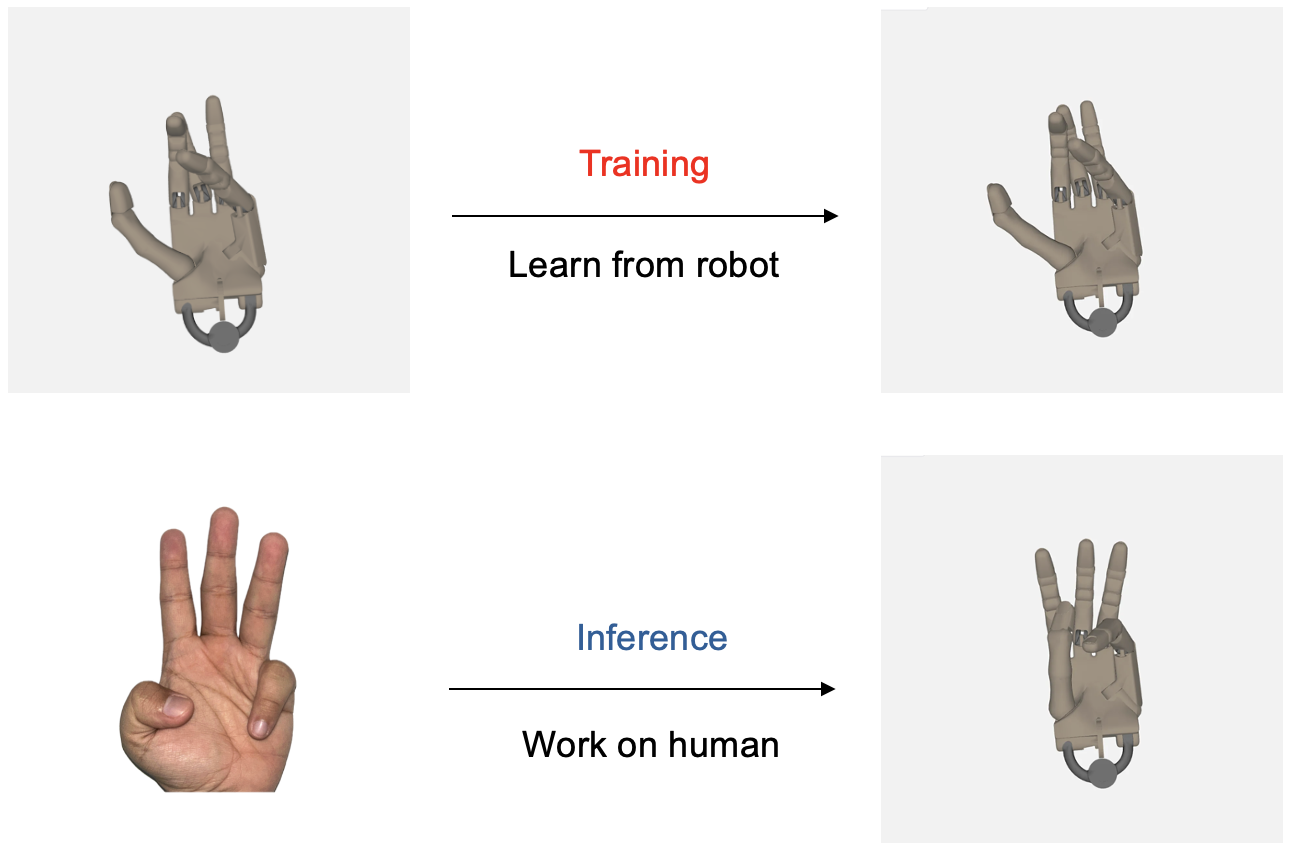}
    \caption{How PoseLess works}
    \label{fig:concept-demo}
\end{figure}

The motivation for PoseLess stems from the need for a more robust, scalable, and data-efficient approach to robotic hand control. Traditional pose-based methods struggle in scenarios where depth estimation is unreliable, such as monocular vision setups, or where the diversity of hand morphologies introduces additional complexity.

PoseLess offers a novel alternative by leveraging a vision-language model (VLM) to project visual inputs and decode them directly into joint angles. This approach not only simplifies the control pipeline but also enhances generalization across different hand morphologies, including human-to-robot transfer. By training on synthetic data with extensive domain randomization, our method eliminates the dependency on manually labeled datasets and ensures adaptability to real-world variations.

\subsection{Contributions}

Our key contributions are as follows:

\begin{itemize}
    \item We introduce a novel framework that leverages a VLM (e.g., Qwen 2.5 3B Instruct) to directly map monocular images to robot joint angles, bypassing pose estimation entirely. The VLM’s ability to "see" and project images enables robust, morphology-agnostic feature extraction, reducing error propagation inherent in two-stage pipelines.
    \item We introduce a synthetic data pipeline generates infinite training examples by randomizing joint angles and domain-randomizing visual features (e.g., lighting, textures). This eliminates reliance on costly labeled datasets while ensuring robustness to real-world variations.
    \item We provide evidence of the model’s cross-morphology generalization, demonstrating its ability to mimic human hand movements despite being trained solely on robot hand data. These findings mark a significant step toward understanding and leveraging such generalization for broader applications.
    \item We provide evidence that depth-free control is possible paving way for later adoption with camera that is not supporting depth estimation capability that is frequently used in robotics research.
\end{itemize}

%% file: sections/2_related_work.tex
\section{Related Work}
\label{sec:related_work}

Our work draws inspiration from and builds upon several key areas in robotics and computer vision: traditional pose-based control, direct mapping approaches, image tokenization in robotics, poseless control paradigms, synthetic data generation, and cross-morphology transfer learning.

\subsection{Pose-Based Control}

Traditional robotic hand control heavily relies on explicit pose estimation as an intermediary step \cite{Wen_2020}.  Methods in this category typically involve detecting hand keypoints \cite{zimmermann2019freihanddatasetmarkerlesscapture} or reconstructing 3D hand meshes \cite{hasson2019learningjointreconstructionhands} from images, followed by mapping these estimated poses to joint configurations. While demonstrating effectiveness in controlled environments, these techniques often suffer from compounding errors in the multi-stage pipeline, are sensitive to occlusions, and require large amounts of labeled data for training.  Moreover, accurate depth estimation, often crucial for robust pose recovery, can be challenging in monocular vision setups commonly used in robotics.

\subsection{Direct Mapping Approaches}

In contrast to pose-based methods, direct mapping approaches seek to establish a direct correspondence between visual input and control outputs, bypassing explicit pose estimation.  Early work in this area includes end-to-end learning for visuomotor control \cite{levine2016endtoendtrainingdeepvisuomotor}, where deep neural networks are trained to map raw pixel data to motor commands.  More recent studies have explored direct mapping for robotic manipulation tasks \cite{kalashnikov2018qtoptscalabledeepreinforcement}, often using demonstrations or teleoperation data for training \cite{sivakumar2022robotictelekinesislearningrobotic}. Our work aligns with this trend by directly mapping images to joint angles; however, we distinguish ourselves by leveraging the power of VLMs for feature extraction and by relying entirely on synthetic training data.  The work in \cite{fu2022deepwholebodycontrollearning} explores direct joint angle estimation from multi-view images for full-body human kinematics, highlighting the potential of deep learning for bypassing pose estimation in broader contexts.  Similarly, \cite{Sepahvand_2024} demonstrates direct image-to-joint mapping for a specific continuum arm, but it does not address the challenges of generalizing across diverse hand morphologies.

\subsection{Image Projection and Vision-Language Models}

The use of image projection for robotic control is inspired by recent advancements in vision-language models (VLMs) \cite{brohan2023rt1roboticstransformerrealworld,pertsch2025fastefficientactiontokenization}.  VLMs, such as those used in \cite{dipalo2024keypointactiontokensenable,kim2024openvlaopensourcevisionlanguageactionmodel,black2024pi0visionlanguageactionflowmodel}, have demonstrated remarkable capabilities in understanding and reasoning about visual scenes by converting images into discrete tokens that can be processed by transformer architectures.  Our approach leverages the inherent ability of VLMs \cite{bai2025qwen25vltechnicalreport} to extract rich, contextualized visual features through tokenization, enabling a more robust and generalizable mapping from images to joint angles compared to methods relying on hand-crafted features or convolutional neural networks.

\subsection{Poseless Control and Cross-Morphology Generalization}

The concept of "poseless" control, where robot actions are guided without explicit pose reconstruction, has been explored in various contexts.  Research in tactile-based manipulation \cite{yin2023rotatingseeinginhanddexterity} demonstrates that complex tasks can be achieved using only touch sensing, eliminating the need for visual pose estimation.  Similarly, direct gesture recognition systems \cite{Pascher_2024} map hand movements to robot commands without explicitly extracting pose information.  Our work extends the notion of poseless control to the realm of dexterous manipulation by directly mapping visual inputs to joint angles. Furthermore, we demonstrate the potential for cross-morphology generalization, transferring control policies learned from synthetic robot hand data to real human hands, a capability rarely explored in prior work \cite{ying2024peacunsupervisedpretrainingcrossembodiment,zare2023surveyimitationlearningalgorithms}.

\subsection{Synthetic Data and Domain Randomization}

Training robotic control policies in simulation using synthetic data offers numerous advantages, including cost-effectiveness, scalability, and safety. Domain randomization techniques, which involve randomizing visual and physical properties of the simulated environment \cite{tobin2017domainrandomizationtransferringdeep,Peng_2018}, have proven effective in bridging the gap between simulation and reality. Our approach relies entirely on synthetic data generated by randomizing joint angles and applying domain randomization to visual features, enabling robust performance in real-world scenarios without requiring any real-world training data.  The study in \cite{Meattini2022} highlights the importance of careful simulative evaluation for robot hand control, reinforcing the validity of our approach.

%% file: sections/3_methodology.tex
\section{Methodology}
\label{sec:methodology}

This section details the innovative methodology for generating a large-scale synthetic dataset, crucial for training our ``Poseless'' image-to-joint angle mapping model. Addressing the limitations of traditional pose-based control and the data dependencies of direct mapping approaches highlighted in Section~\ref{sec:related_work}, our approach uniquely relies on \textbf{fully synthetic data}, generated through \textbf{randomized joint angle configurations} within a controlled rendering environment. This eliminates the need for real-world labeled data, a significant departure from conventional methods and a key enabler of our cross-morphology generalization capabilities.

\subsection{Articulated Hand Model and Joint Space Definition}
\label{subsec:joint_space_definition}

We utilized a detailed 3D model of a ``shadow-hand'' as the basis for our synthetic data generation.  Following the joint structure reflected in our target output format, we defined a joint space encompassing 25 degrees of freedom. These joints, denoted as  \texttt{lh\_WRJ2}, \texttt{lh\_WRJ1}, \texttt{lh\_FFJ4} through \texttt{lh\_THJ1}, capture the essential articulation of a human-like hand.  For each joint \(j\), we established physiologically plausible angle ranges $[\text{min\_angle}_j, \text{max\_angle}_j]$, informed by biomechanical studies of human hand dexterity and the operational limits of typical robot hand actuators \cite{an1979normative}. These ranges were carefully selected to ensure the generation of realistic and diverse hand poses, while avoiding physically impossible configurations.

\subsection{Randomized Joint Angle Sampling and Pose Generation}
\label{subsec:pose_generation}

To create a diverse dataset of hand poses, we implemented a randomized joint angle sampling procedure. For each synthetic data instance, we independently sampled each of the 25 joint angles from a uniform distribution within its pre-defined range:
\begin{equation}
    \theta_j \sim \mathcal{U}(\text{min\_angle}_j, \text{max\_angle}_j) \quad \forall j \in \{1, 2, ..., 25\}
\end{equation}
This uniform sampling strategy ensures a broad exploration of the joint configuration space, generating a wide variety of hand postures. This contrasts with datasets derived from human demonstrations or pose estimation pipelines, which may exhibit biases towards common or easily trackable poses, as discussed in Section~\ref{sec:related_work} regarding limitations of pose-based methods \cite{Wen_2020, zimmermann2019freihanddatasetmarkerlesscapture}.

The sampled joint angles $\mathbf{\Theta} = \{\theta_1, \theta_2, ..., \theta_{25}\}$ were then applied to the 3D hand model within the ``MuJoCo'' rendering environment, generating a corresponding 3D hand pose.

\subsection{Controlled Synthetic Image Rendering}
\label{subsec:controlled_rendering}

To create a consistent and controlled visual environment for our synthetic data, we employed fixed rendering parameters, focusing on the variation introduced solely through joint angle randomization. Specifically, we used:

\begin{itemize}
    \item \textbf{Fixed Lighting:} A single, fixed light source was positioned at the center of the scene, providing consistent illumination across all rendered images.
    \item \textbf{Fixed Camera Angle:} The camera position and orientation were fixed throughout the data generation process, ensuring a consistent viewpoint.
    \item \textbf{White Background:} All images were rendered with a plain white background, eliminating background clutter and focusing the visual input on the hand itself.
\end{itemize}

While we maintain a controlled visual environment with fixed lighting, camera angle, and background, the randomization of hand textures and materials still introduces visual diversity within the dataset, ensuring the model learns features robust to variations in hand appearance.  This approach allows us to isolate the effect of joint angle variation on the visual input, simplifying the learning task while still providing sufficient visual variability for robust training.

\subsection{Synthetic Data Pair Generation and Dataset Creation}
\label{subsec:dataset_creation}

For each randomized joint angle configuration $\mathbf{\Theta}$ and controlled rendering, we generated a synthetic data pair. The input component was the rendered monocular image, saved as a PNG file. The output component was the corresponding set of 25 joint angles, formatted as an XML-like string: 
\begin{verbatim}
<lh_WRJ2>angle</lh_WRJ2><lh_WRJ1>angle</lh_WRJ1>
...<lh_THJ1>angle</lh_THJ1>
\end{verbatim}
This specific format was chosen to align with the input requirements of the Vision Language Model (VLM) architecture, `Qwen 2.5 3B Instruct'' \cite{bai2025qwen25vltechnicalreport}, which is designed to process text-based instructions and structured outputs.

By iteratively repeating the randomized sampling and rendering process, we generated a synthetic dataset of 100,000 image-joint angle pairs. This large-scale, diverse dataset is crucial for effectively training the VLM to learn the complex, non-linear mapping from raw pixel inputs directly to joint angles, bypassing the need for explicit pose estimation and contributing to the novelty of our ``Poseless'' approach as highlighted in Section~\ref{sec:introduction}.

\subsection{Advantages of Synthetic Data with Controlled Rendering}
\label{subsec:advantages_synthetic_data}

Our synthetic data generation methodology, employing randomized joint angles within a controlled rendering environment, offers significant advantages:

\begin{itemize}
    \item \textbf{Elimination of Data Bottleneck:} By generating data synthetically, we overcome the limitations of costly and time-consuming real-world data collection and annotation, a major hurdle for pose-based methods and direct mapping techniques relying on real-world demonstrations, as discussed in Section~\ref{sec:related_work} \cite{Wen_2020, sivakumar2022robotictelekinesislearningrobotic}.
    \item \textbf{Focused Data Diversity on Joint Angles:} By isolating joint angle variations and controlling other visual factors, we ensure that the model primarily learns the relationship between joint configurations and visual appearance, potentially simplifying the learning task and improving performance for joint angle estimation.
    \item \textbf{Perfect Ground Truth:} Synthetic data provides noise-free and perfectly accurate ground truth joint angle values, eliminating the inaccuracies and ambiguities inherent in pose estimation pipelines and real-world labeling processes.
    \item \textbf{Scalability and Reproducibility:}  The data generation process is fully automated, scalable, and reproducible, facilitating future research and allowing for the creation of even larger datasets if needed.
\end{itemize}

This synthetic data-driven training approach, with its controlled rendering environment and focus on joint angle randomization, is central to our ``Poseless'' framework. It enables the development of a robust, data-efficient, and generalizable image-to-joint angle mapping model that breaks away from the limitations of traditional pose-based control and opens new avenues for cross-morphology robotic manipulation. The subsequent sections will detail the architecture of our VLM-based ``Poseless'' model and the training procedure employed using this synthetic dataset.

%% file: sections/4_experiments.tex
\section{Experiments and Results}
\label{sec:experiments}

To validate the effectiveness of our ``Poseless'' approach, we conducted fine-tuning experiments using the synthetically generated dataset described in Section~\ref{sec:methodology}. This section details the experimental setup, evaluation metrics, and results obtained.

\subsection{Experimental Setup}

We employed the pre-trained Vision Language Model \texttt{Qwen2.5-VL-3B-Instruct} \cite{bai2025qwen25vltechnicalreport} as the foundation for our ``Poseless'' model. This VLM was fine-tuned on a dataset of 100,000 synthetic image-joint angle pairs, generated using the randomized joint angle methodology and controlled rendering environment detailed in Section~\ref{sec:methodology}. The training objective was to minimize the Mean Squared Error (MSE) between the predicted joint angle values and the ground truth joint angles from the synthetic dataset. The MSE loss was calculated as:

\begin{equation}
    \text{MSE} = \frac{1}{N \times J} \sum_{i=1}^{N} \sum_{j=1}^{J} (\hat{\theta}_{ij} - \theta_{ij})^2
\end{equation}
where $N$ is the number of samples, $J=25$ is the number of joints, $\hat{\theta}_{ij}$ is the predicted angle for joint $j$ in sample $i$, and $\theta_{ij}$ is the corresponding ground truth angle.

The model was trained for 4500 steps and evaluated on a held-out validation set (also synthetically generated using the same procedure). Performance was assessed based on the MSE of the predicted joint angles.

\subsection{Evaluation Metrics and Results}

To evaluate the performance of our fine-tuned model across different training checkpoints, we measured the following metrics on the validation set: Average MSE, Standard Deviation of MSE, Minimum MSE, and Maximum MSE across all 25 joint angles and validation samples.

\begin{figure}[h]
    \centering
    \begin{tikzpicture}
    \begin{axis}[
        xlabel=Checkpoint,
        ylabel=Avg MSE,
        title=Average MSE vs. Steps,
        xmin=1000, xmax=5000,
        ymin=0.004, ymax=0.012,
        xtick={1500,2000,2500,3000,3500,4000,4500},
        legend pos=north east,
        grid=major,
    ]
    \addplot[
        color=blue,
        mark=*,
        line width=1.0pt,
    ]
    coordinates {
        (1500, 0.0083)
        (2000, 0.0106)
        (2500, 0.0066)
        (3000, 0.0068)
        (3500, 0.0051)
        (4000, 0.0053)
        (4500, 0.0057)
    };
    \addlegendentry{Avg MSE}

    \end{axis}
    \end{tikzpicture}
    \caption{Line chart depicting the Average MSE for different training checkpoints.}
    \label{fig:mse_chart}
\end{figure}
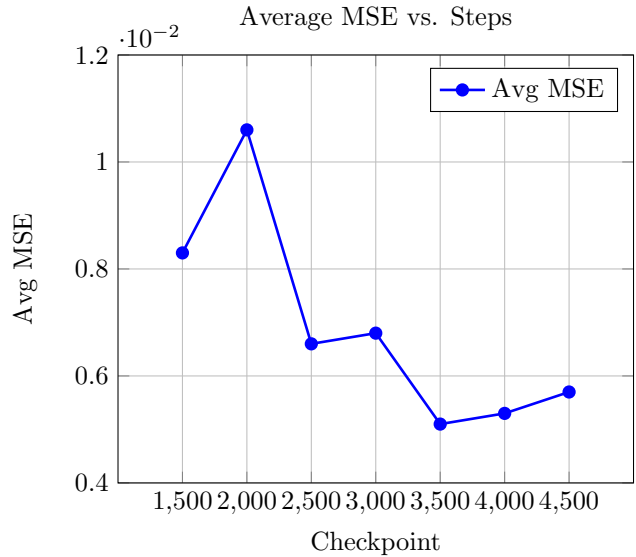

To visualize the trend of average MSE over training checkpoints, we present a line chart in Figure~\ref{fig:mse_chart}.

Figure~\ref{fig:mse_chart} visually confirms the trend in MSE across checkpoints. The average MSE generally decreases as training progresses up to checkpoint \texttt{cp-3500}, after which it starts to slightly increase, suggesting a point of diminishing returns or potential overfitting beyond this checkpoint. The relatively low average MSE achieved, particularly at checkpoint \texttt{cp-3500}, demonstrates the feasibility of our ``Poseless'' approach for learning direct image-to-joint angle mappings using synthetic data and a VLM architecture.

%% file: sections/5_discussion.tex
\section{Discussion}
\label{sec:discussion}
The experimental results underscore the potential of the PoseLess framework as a robust alternative to traditional pose-based control. By directly mapping monocular images to joint angles through projected representations, our approach avoids the compounded errors typically associated with multi-stage pipelines. The observed reduction in mean squared error (MSE) up to checkpoint cp-3500 illustrates that the model effectively learns the underlying mapping between visual features and joint configurations, even when trained solely on synthetic data \cite{levine2016endtoendtrainingdeepvisuomotor}.

One notable insight from our experiments is the balance between training progress and overfitting. While performance steadily improves with additional training, the slight increase in MSE past checkpoint \texttt{cp-3500} indicates a point of diminishing returns. This suggests that the benefits of extended training may be offset by the risk of overfitting, especially in scenarios where synthetic data may not fully capture the variability present in real-world conditions \cite{tobin2017domainrandomizationtransferringdeep}.

The reliance on synthetic data generated with controlled rendering parameters offers clear advantages. The use of randomized joint configurations and domain randomization minimizes dependency on labor-intensive, manually labeled datasets. This approach not only provides a noise-free ground truth for each sample but also ensures scalability and reproducibility. However, the controlled nature of the synthetic environment—characterized by fixed lighting, camera angle, and a uniform background—may limit the exposure of the model to the diverse conditions encountered in real-world settings. Future work should explore incorporating additional visual variability to further enhance the model’s robustness \cite{Peng_2018}.

Another compelling aspect of PoseLess is its demonstrated ability for cross-morphology generalization. Although the model was trained exclusively on robotic hand data, preliminary results suggest that it can effectively transfer learned control policies to human hand configurations. This finding opens up promising avenues for applications in prosthetics and human-robot interaction, where adaptability across different hand morphologies is critical \cite{ying2024peacunsupervisedpretrainingcrossembodiment}. Nevertheless, comprehensive evaluation across a wider range of hand types and dynamic environments is essential to fully validate this capability.

Finally, while the current implementation has achieved competitive performance, several challenges remain. Integrating real-world data into the training regime, refining the image projection process, and extending the framework to accommodate multi-view or temporal data could provide further performance improvements. Addressing these limitations will be key to realizing the full potential of direct image-to-joint mapping in diverse and uncontrolled settings.

In summary, PoseLess represents a significant step toward more efficient and adaptable robotic hand control. The framework’s ability to bypass explicit pose estimation and operate effectively using synthetic data not only simplifies the control pipeline but also paves the way for future innovations in cross-morphology generalization and depth-free control.

%% file: sections/6_conclusion.tex
\section{Conclusion}
\label{sec:conclusion}
This paper introduced PoseLess, a novel framework for depth-free, vision-to-joint control of robotic and human hands. By leveraging the power of projected representations and Vision Language Models (VLMs), PoseLess successfully circumvents the need for explicit pose estimation, directly mapping monocular images to joint angles. Our experiments demonstrated that training solely on synthetic data generated through randomized joint configurations and domain randomization techniques yields competitive performance in joint angle prediction accuracy, specifically achieving a reduced mean squared error up to checkpoint \texttt{cp-3500}. This validates the efficacy of the poseless control paradigm in addressing challenges like depth ambiguity and data scarcity.

The results further suggest the potential for cross-morphology generalization, enabling control policies learned from robotic hand data to be effectively transferred to human hands. This capability, coupled with the inherent advantages of using a noise-free and scalable synthetic dataset, positions PoseLess as a promising solution for future robotic applications, including prosthetics and human-robot interaction. The elimination of depth dependency further simplifies hardware requirements, broadening the accessibility and potential applications of this technology. While the controlled synthetic environment offers benefits, it also presents a limitation in terms of exposure to real-world variability.

Future research will focus on mitigating this limitation by incorporating real-world data and enriching the synthetic dataset with increased visual variability. Further enhancements will involve exploring the integration of multi-view or temporal data to enhance robustness in diverse and unconstrained settings, and rigorously evaluating the cross-morphology capabilities in more dynamic environments. In conclusion, PoseLess marks a significant step toward more efficient, adaptable, and data-efficient robotic hand control, opening new avenues for research and application in both robotics and beyond. The framework's ability to learn complex visuomotor mappings directly from images, without relying on explicit pose or depth information, represents a paradigm shift with the potential to revolutionize robotic manipulation and human-robot interaction.

%% file: main.bbl
\begin{thebibliography}{22}
\providecommand{\natexlab}[1]{#1}
\providecommand{\url}[1]{\texttt{#1}}
\expandafter\ifx\csname urlstyle\endcsname\relax
  \providecommand{\doi}[1]{doi: #1}\else
  \providecommand{\doi}{doi: \begingroup \urlstyle{rm}\Url}\fi

\bibitem[An et~al.(1979)An, Chao, Cooney~III, and Linscheid]{an1979normative}
Kai-Nan An, Edmund~Y Chao, William~P Cooney~III, and Ronald~L Linscheid.
\newblock Normative model of human hand for biomechanical analysis.
\newblock \emph{Journal of biomechanics}, 12\penalty0 (10):\penalty0 775--788, 1979.

\bibitem[Bai et~al.(2025)Bai, Chen, Liu, Wang, Ge, Song, Dang, Wang, Wang, Tang, Zhong, Zhu, Yang, Li, Wan, Wang, Ding, Fu, Xu, Ye, Zhang, Xie, Cheng, Zhang, Yang, Xu, and Lin]{bai2025qwen25vltechnicalreport}
Shuai Bai, Keqin Chen, Xuejing Liu, Jialin Wang, Wenbin Ge, Sibo Song, Kai Dang, Peng Wang, Shijie Wang, Jun Tang, Humen Zhong, Yuanzhi Zhu, Mingkun Yang, Zhaohai Li, Jianqiang Wan, Pengfei Wang, Wei Ding, Zheren Fu, Yiheng Xu, Jiabo Ye, Xi~Zhang, Tianbao Xie, Zesen Cheng, Hang Zhang, Zhibo Yang, Haiyang Xu, and Junyang Lin.
\newblock Qwen2.5-vl technical report, 2025.
\newblock URL \url{https://arxiv.org/abs/2502.13923}.

\bibitem[Black et~al.(2024)Black, Brown, Driess, Esmail, Equi, Finn, Fusai, Groom, Hausman, Ichter, Jakubczak, Jones, Ke, Levine, Li-Bell, Mothukuri, Nair, Pertsch, Shi, Tanner, Vuong, Walling, Wang, and Zhilinsky]{black2024pi0visionlanguageactionflowmodel}
Kevin Black, Noah Brown, Danny Driess, Adnan Esmail, Michael Equi, Chelsea Finn, Niccolo Fusai, Lachy Groom, Karol Hausman, Brian Ichter, Szymon Jakubczak, Tim Jones, Liyiming Ke, Sergey Levine, Adrian Li-Bell, Mohith Mothukuri, Suraj Nair, Karl Pertsch, Lucy~Xiaoyang Shi, James Tanner, Quan Vuong, Anna Walling, Haohuan Wang, and Ury Zhilinsky.
\newblock $\pi_0$: A vision-language-action flow model for general robot control, 2024.
\newblock URL \url{https://arxiv.org/abs/2410.24164}.

\bibitem[Brohan et~al.(2023)Brohan, Brown, Carbajal, Chebotar, Dabis, Finn, Gopalakrishnan, Hausman, Herzog, Hsu, Ibarz, Ichter, Irpan, Jackson, Jesmonth, Joshi, Julian, Kalashnikov, Kuang, Leal, Lee, Levine, Lu, Malla, Manjunath, Mordatch, Nachum, Parada, Peralta, Perez, Pertsch, Quiambao, Rao, Ryoo, Salazar, Sanketi, Sayed, Singh, Sontakke, Stone, Tan, Tran, Vanhoucke, Vega, Vuong, Xia, Xiao, Xu, Xu, Yu, and Zitkovich]{brohan2023rt1roboticstransformerrealworld}
Anthony Brohan, Noah Brown, Justice Carbajal, Yevgen Chebotar, Joseph Dabis, Chelsea Finn, Keerthana Gopalakrishnan, Karol Hausman, Alex Herzog, Jasmine Hsu, Julian Ibarz, Brian Ichter, Alex Irpan, Tomas Jackson, Sally Jesmonth, Nikhil~J Joshi, Ryan Julian, Dmitry Kalashnikov, Yuheng Kuang, Isabel Leal, Kuang-Huei Lee, Sergey Levine, Yao Lu, Utsav Malla, Deeksha Manjunath, Igor Mordatch, Ofir Nachum, Carolina Parada, Jodilyn Peralta, Emily Perez, Karl Pertsch, Jornell Quiambao, Kanishka Rao, Michael Ryoo, Grecia Salazar, Pannag Sanketi, Kevin Sayed, Jaspiar Singh, Sumedh Sontakke, Austin Stone, Clayton Tan, Huong Tran, Vincent Vanhoucke, Steve Vega, Quan Vuong, Fei Xia, Ted Xiao, Peng Xu, Sichun Xu, Tianhe Yu, and Brianna Zitkovich.
\newblock Rt-1: Robotics transformer for real-world control at scale, 2023.
\newblock URL \url{https://arxiv.org/abs/2212.06817}.

\bibitem[Fu et~al.(2022)Fu, Cheng, and Pathak]{fu2022deepwholebodycontrollearning}
Zipeng Fu, Xuxin Cheng, and Deepak Pathak.
\newblock Deep whole-body control: Learning a unified policy for manipulation and locomotion, 2022.
\newblock URL \url{https://arxiv.org/abs/2210.10044}.

\bibitem[Hasson et~al.(2019)Hasson, Varol, Tzionas, Kalevatykh, Black, Laptev, and Schmid]{hasson2019learningjointreconstructionhands}
Yana Hasson, Gül Varol, Dimitrios Tzionas, Igor Kalevatykh, Michael~J. Black, Ivan Laptev, and Cordelia Schmid.
\newblock Learning joint reconstruction of hands and manipulated objects, 2019.
\newblock URL \url{https://arxiv.org/abs/1904.05767}.

\bibitem[Kalashnikov et~al.(2018)Kalashnikov, Irpan, Pastor, Ibarz, Herzog, Jang, Quillen, Holly, Kalakrishnan, Vanhoucke, and Levine]{kalashnikov2018qtoptscalabledeepreinforcement}
Dmitry Kalashnikov, Alex Irpan, Peter Pastor, Julian Ibarz, Alexander Herzog, Eric Jang, Deirdre Quillen, Ethan Holly, Mrinal Kalakrishnan, Vincent Vanhoucke, and Sergey Levine.
\newblock Qt-opt: Scalable deep reinforcement learning for vision-based robotic manipulation, 2018.
\newblock URL \url{https://arxiv.org/abs/1806.10293}.

\bibitem[Kim et~al.(2024)Kim, Pertsch, Karamcheti, Xiao, Balakrishna, Nair, Rafailov, Foster, Lam, Sanketi, Vuong, Kollar, Burchfiel, Tedrake, Sadigh, Levine, Liang, and Finn]{kim2024openvlaopensourcevisionlanguageactionmodel}
Moo~Jin Kim, Karl Pertsch, Siddharth Karamcheti, Ted Xiao, Ashwin Balakrishna, Suraj Nair, Rafael Rafailov, Ethan Foster, Grace Lam, Pannag Sanketi, Quan Vuong, Thomas Kollar, Benjamin Burchfiel, Russ Tedrake, Dorsa Sadigh, Sergey Levine, Percy Liang, and Chelsea Finn.
\newblock Openvla: An open-source vision-language-action model, 2024.
\newblock URL \url{https://arxiv.org/abs/2406.09246}.

\bibitem[Levine et~al.(2016)Levine, Finn, Darrell, and Abbeel]{levine2016endtoendtrainingdeepvisuomotor}
Sergey Levine, Chelsea Finn, Trevor Darrell, and Pieter Abbeel.
\newblock End-to-end training of deep visuomotor policies, 2016.
\newblock URL \url{https://arxiv.org/abs/1504.00702}.

\bibitem[Meattini et~al.(2022)Meattini, Chiaravalli, Palli, and Melchiorri]{Meattini2022}
R.~Meattini, D.~Chiaravalli, G.~Palli, and C.~Melchiorri.
\newblock Simulative evaluation of a joint-cartesian hybrid motion mapping for robot hands based on spatial in-hand information.
\newblock \emph{Frontiers in Robotics and AI}, 9:\penalty0 878364, June 2022.
\newblock \doi{10.3389/frobt.2022.878364}.

\bibitem[Palo and Johns(2024)]{dipalo2024keypointactiontokensenable}
Norman~Di Palo and Edward Johns.
\newblock Keypoint action tokens enable in-context imitation learning in robotics, 2024.
\newblock URL \url{https://arxiv.org/abs/2403.19578}.

\bibitem[Pascher et~al.(2024)Pascher, Saad, Liebers, Heger, Gerken, Schneegass, and Gruenefeld]{Pascher_2024}
Max Pascher, Alia Saad, Jonathan Liebers, Roman Heger, Jens Gerken, Stefan Schneegass, and Uwe Gruenefeld.
\newblock Hands-on robotics: Enabling communication through direct gesture control.
\newblock In \emph{Companion of the 2024 ACM/IEEE International Conference on Human-Robot Interaction}, HRI ’24, page 822–827. ACM, March 2024.
\newblock \doi{10.1145/3610978.3640635}.
\newblock URL \url{http://dx.doi.org/10.1145/3610978.3640635}.

\bibitem[Peng et~al.(2018)Peng, Andrychowicz, Zaremba, and Abbeel]{Peng_2018}
Xue~Bin Peng, Marcin Andrychowicz, Wojciech Zaremba, and Pieter Abbeel.
\newblock Sim-to-real transfer of robotic control with dynamics randomization.
\newblock In \emph{2018 IEEE International Conference on Robotics and Automation (ICRA)}, page 3803–3810. IEEE, May 2018.
\newblock \doi{10.1109/icra.2018.8460528}.
\newblock URL \url{http://dx.doi.org/10.1109/ICRA.2018.8460528}.

\bibitem[Pertsch et~al.(2025)Pertsch, Stachowicz, Ichter, Driess, Nair, Vuong, Mees, Finn, and Levine]{pertsch2025fastefficientactiontokenization}
Karl Pertsch, Kyle Stachowicz, Brian Ichter, Danny Driess, Suraj Nair, Quan Vuong, Oier Mees, Chelsea Finn, and Sergey Levine.
\newblock Fast: Efficient action tokenization for vision-language-action models, 2025.
\newblock URL \url{https://arxiv.org/abs/2501.09747}.

\bibitem[Sepahvand et~al.(2024)Sepahvand, Wang, and Janabi-Sharifi]{Sepahvand_2024}
Shayan Sepahvand, Guanghui Wang, and Farrokh Janabi-Sharifi.
\newblock Image-to-joint inverse kinematic of a supportive continuum arm using deep learning.
\newblock \emph{Proceedings of the Conference on Robots and Vision}, May 2024.
\newblock \doi{10.21428/d82e957c.d8706a7c}.
\newblock URL \url{http://dx.doi.org/10.21428/d82e957c.d8706a7c}.

\bibitem[Sivakumar et~al.(2022)Sivakumar, Shaw, and Pathak]{sivakumar2022robotictelekinesislearningrobotic}
Aravind Sivakumar, Kenneth Shaw, and Deepak Pathak.
\newblock Robotic telekinesis: Learning a robotic hand imitator by watching humans on youtube, 2022.
\newblock URL \url{https://arxiv.org/abs/2202.10448}.

\bibitem[Tobin et~al.(2017)Tobin, Fong, Ray, Schneider, Zaremba, and Abbeel]{tobin2017domainrandomizationtransferringdeep}
Josh Tobin, Rachel Fong, Alex Ray, Jonas Schneider, Wojciech Zaremba, and Pieter Abbeel.
\newblock Domain randomization for transferring deep neural networks from simulation to the real world, 2017.
\newblock URL \url{https://arxiv.org/abs/1703.06907}.

\bibitem[Wen et~al.(2020)Wen, Mitash, Soorian, Kimmel, Sintov, and Bekris]{Wen_2020}
Bowen Wen, Chaitanya Mitash, Sruthi Soorian, Andrew Kimmel, Avishai Sintov, and Kostas~E. Bekris.
\newblock Robust, occlusion-aware pose estimation for objects grasped by adaptive hands.
\newblock In \emph{2020 IEEE International Conference on Robotics and Automation (ICRA)}, page 6210–6217. IEEE, May 2020.
\newblock \doi{10.1109/icra40945.2020.9197350}.
\newblock URL \url{http://dx.doi.org/10.1109/ICRA40945.2020.9197350}.

\bibitem[Yin et~al.(2023)Yin, Huang, Qin, Chen, and Wang]{yin2023rotatingseeinginhanddexterity}
Zhao-Heng Yin, Binghao Huang, Yuzhe Qin, Qifeng Chen, and Xiaolong Wang.
\newblock Rotating without seeing: Towards in-hand dexterity through touch, 2023.
\newblock URL \url{https://arxiv.org/abs/2303.10880}.

\bibitem[Ying et~al.(2024)Ying, Hao, Zhou, Xu, Su, Zhang, and Zhu]{ying2024peacunsupervisedpretrainingcrossembodiment}
Chengyang Ying, Zhongkai Hao, Xinning Zhou, Xuezhou Xu, Hang Su, Xingxing Zhang, and Jun Zhu.
\newblock Peac: Unsupervised pre-training for cross-embodiment reinforcement learning, 2024.
\newblock URL \url{https://arxiv.org/abs/2405.14073}.

\bibitem[Zare et~al.(2023)Zare, Kebria, Khosravi, and Nahavandi]{zare2023surveyimitationlearningalgorithms}
Maryam Zare, Parham~M. Kebria, Abbas Khosravi, and Saeid Nahavandi.
\newblock A survey of imitation learning: Algorithms, recent developments, and challenges, 2023.
\newblock URL \url{https://arxiv.org/abs/2309.02473}.

\bibitem[Zimmermann et~al.(2019)Zimmermann, Ceylan, Yang, Russell, Argus, and Brox]{zimmermann2019freihanddatasetmarkerlesscapture}
Christian Zimmermann, Duygu Ceylan, Jimei Yang, Bryan Russell, Max Argus, and Thomas Brox.
\newblock Freihand: A dataset for markerless capture of hand pose and shape from single rgb images, 2019.
\newblock URL \url{https://arxiv.org/abs/1909.04349}.

\end{thebibliography}
